\documentclass[10pt, conference, letterpaper]{IEEEtran}
\IEEEoverridecommandlockouts
\usepackage{cite}
\usepackage{amsmath,amssymb,amsfonts}
\usepackage{graphicx}
\usepackage{textcomp}
\usepackage{xcolor}
\usepackage{comment}
\graphicspath{{SECON_2023_figures/}{current_draft_figures/}{figures/}}

\usepackage{mathtools}
\usepackage{float}
\usepackage{algorithm}
\usepackage{algorithmicx}
\usepackage{algpseudocode}
\algtext*{EndIf}
\algtext*{EndFor}
\algtext*{EndWhile}
\algtext*{EndFunction}
\algtext*{EndProcedure}

\def\BibTeX{{\rm B\kern-.05em{\sc i\kern-.025em b}\kern-.08em
    T\kern-.1667em\lower.7ex\hbox{E}\kern-.125emX}}

\usepackage{tikz}
\newcommand\copyrighttext{%
  \footnotesize \textcopyright 2025 IEEE. Personal use of this material is permitted. 
  Permission from IEEE must be obtained for all other uses, in any current or future 
  media, including reprinting/republishing this material for advertising or 
  promotional purposes, creating new collective works, for resale or redistribution 
  to servers or lists, or reuse of any copyrighted component of this work in other works. Published in: 2024 IEEE/ACM Ninth International Conference on Internet-of-Things Design and Implementation (IoTDI). 
  DOI: 10.1109/IoTDI61053.2024.00014}
\newcommand\copyrightnotice{%
\begin{tikzpicture}[remember picture,overlay]
\node[anchor=south,yshift=10pt] at (current page.south) 
{\fbox{\parbox{\dimexpr\textwidth-\fboxsep-\fboxrule\relax}{\copyrighttext}}};
\end{tikzpicture}%
}

\begin{document}

\title{NaviSlim: Adaptive Context-Aware Navigation and Sensing via Dynamic Slimmable Networks
}

\author{\IEEEauthorblockN{1\textsuperscript{st} Timothy K Johnsen}
\IEEEauthorblockA{\textit{Computational Sciences, Joint Doctoral Program} \\ \textit{University of California, Irvine, USA} \\ \textit{San Diego State University, USA} \\
tjohnsen@uci.edu}
\and
\IEEEauthorblockN{2\textsuperscript{nd} Marco Levorato}
\IEEEauthorblockA{\textit{Computer Science Department} \\
\textit{University of California, Irvine, USA}\\
levorato@uci.edu}}

\maketitle
\copyrightnotice

\begin{abstract}
Small-scale autonomous airborne vehicles, such as micro-drones, are expected to be a central component of a broad spectrum of applications ranging from exploration to surveillance and delivery. This class of vehicles is characterized by severe constraints in computing power and energy reservoir, which impairs their ability to support the complex state-of-the-art neural models needed for autonomous operations. The main contribution of this paper is a new class of neural navigation models -- NaviSlim -- capable of adapting the amount of resources spent on computing and sensing in response to the current context (\textit{i.e.}, difficulty of the environment, current trajectory, and navigation goals). Specifically, NaviSlim is designed as a gated slimmable neural network architecture that, different from existing slimmable networks, can dynamically select a slimming factor to autonomously scale model complexity, which consequently optimizes execution time and energy consumption. Moreover, different from existing sensor fusion approaches, NaviSlim can dynamically select power levels of onboard sensors to autonomously reduce power and time spent during sensor acquisition, without the need to switch between different neural networks. By means of extensive training and testing on the robust simulation environment Microsoft AirSim, we evaluate our NaviSlim models on scenarios with varying difficulty and a test set that showed a dynamic reduced model complexity on average between 57-92\%, and between 61-80\% sensor utilization, as compared to static neural networks designed to match computing and sensing of that required by the most difficult scenario.
\end{abstract}


\section{Introduction}
\label{sec:introduction}

Drone autonomy is a rapidly developing area of investigation among Internet of Things (IoT) devices, with potential impact to a broad range of applications such as remote exploration, first response, agriculture, and delivery. An extensive survey on hardware and software requirements for developing fully autonomous Unmanned Aerial Vehicles (UAV) can be found in \cite{b1}. The most significant issues reported by the authors are: increasing complexity of tasks, operating in unknown and diverse environments, limitations on sensing capabilities, flight time, and energy consumption. 

The aforementioned issues become more prominent when considering small-scale vehicles such as airborne micro-drones. This class of vehicles suffers from extreme constraints in terms of computational power and energy reservoir, which severely limits their ability to support the complex machine learning algorithms prevalent in vehicular autonomy. Moreover, the same limitations affect onboard sensors, that require time and energy to achieve adequate resolution to support autonomous functionalities in dynamic environments. Intuitively, the time required to execute sensor acquisition and machine learning algorithms during inference also slow down the reaction time of autonomous vehicles. 

The literature related towards developing (micro-)drone autonomy is rich in autonomous drone racing  \cite{rojas2021board}; while other efforts are in developing autonomous swarms \cite{irizarry2022scalable}, or in robust test-beds to evaluate them in \cite{clark2014autonomous}. General studies towards drone autonomy have focused on reducing computation required to execute static neural models as in \cite{anwar2020autonomous, mehra2020reviewnet}, fusing sensors \cite{yeong2021sensor}, and of course general autonomous navigation logic and modeling such as in \cite{amer2021deep}. We find the literature lacking in studies that place an emphasis on developing autonomous navigation methods specifically for micro-drone systems with extreme resource constraints. To this aim, we contend that it is necessary to evolve the static nature of state-of-the-art navigation models, and neural models in general, into dynamic algorithms that use the minimum amount of computational complexity required by the difficulty of scenario, and concurrently the minimum time and energy spent in sensor acquisition.

To accomplish such an ambitious objective, in this paper we introduce a new class of navigation models that have a dynamic architecture capable of adapting its structure in real-time to the difficulty of the current operating environment at a fine-time granularity. Specifically, we present \emph{NaviSlim}, a navigational neural network architecture that dynamically scales its own complexity and sensor modalities based on the perceived context. To support this logic, we propose a robust design and multi-stage training approach based primarily on slimmable networks \cite{b20}, which train partitioned sub-networks within a larger static one, and knowledge distillation \cite{b32}, which trains different models to exhibit similar behavior. In addition to these two core components, our design utilizes a broad array of advanced algorithms and methods such as shortest path algorithms, supervised learning, deep reinforcement learning, and curriculum learning. 

We develop our models using a test bed environment with a simulation tool that can be found open-source on our GitHub \footnote{https://github.com/WreckItTim/rl\_drone}. Our Python repository interfaces with Microsoft AirSim \cite{b3} which is a robust drone simulator rendered in Unreal Engine \cite{unrealengine} to handle physics and graphics. Our experiments evaluate several scenarios with varying difficulties. We show that our \emph{NaviSlim} models are dynamically reduced, on average, to: 1) 57-92\% model complexity, and 2) 61-80\% sensor utilization, of that used by a static network required to otherwise safely navigate the terrain. We further provide evidence that \emph{NaviSlim} adapts the resources used (power, time, and energy) to perceived context. 

The remainder of this paper is organized as follows. We first present literature related to dynamic neural networks, and identify the gap we aim to fill within autonomous drone navigation. Then we provide an overview of our approach in Section~\ref{sec:overview}, including the general system model design, problem formulation, and inherent challenges. The test bed environment and simulation tool are presented in Section~\ref{sec:environment}. The implementation of \emph{NaviSlim} is described in Section~\ref{sec:impementation}, while the training procedure is discussed in Section~\ref{sec:navigation} and Section~\ref{sec:auxiliary}. We end with results of our experiments implemented in the test bed environment, and conclusions. 

\section{Related Work}
\label{sec:related}

State-of-the-art navigation models employ deep reinforcement learning, as in \cite{b18, b6, b15, b4, b17, b19}, in the form of a static neural model processing data from a fixed sensor array -- which we use as a comparison to our presented models. The core limitation of such state-of-the-art approaches is that the model and sensing characteristics need to match the most challenging operating situation, which leads to an unnecessarily large resource usage -- \textit{i.e.}, the neural network models and sensing requirements are static. Several frameworks are deployed as Simultaneous Localization And Mapping (SLAM) approaches, which includes methods for navigation such as localization, mapping, and tracking \cite{whyte2006simultaneous}. However, SLAM algorithms require extensive sensing and intense computing, and are impractical for microdrones - which are the focus of this paper..

Our proposed architecture falls under the general umbrella of dynamic neural networks, which can scale the depth of a neural network (vertically) with early exits \cite{b32} and the width of the network (horizontally) with slimmable networks \cite{b20}.
The most popular class of dynamic neural models use early exits \cite{b32}, where low complexity structures (the early exits) are attached to a main model and are sequentially executed. The processing of an input is terminated if the output of the last executed exit has sufficient confidence, so that the overall number of operations depends on the complexity of individual input samples. Our architecture belongs to a new emerging class of dynamic neural networks -- dynamic slimmable models -- adopted by a small number of recent contributions~\cite{b23, b33}, where an internal module manipulates characteristics of the entire network.

An alternative approach to ours is to store multiple versions of the same model and swap models at runtime, an approach that requires extended memory availability, as well as a potential context switching latency. Neural architecture search (NAS) \cite{b34} embeds a DRL algorithm to select the optimal network structure, however is extremely time consuming as each iteration requires complete training. In our approach, we develop models that realize an advanced form of dynamic slimmable networks \cite{b20}, designed to seamlessly change its shape with minimal memory and no latency overhead. Technically, this is accomplished by horizontally scaling down the number of active nodes in a set of target layer(s) within a larger super-network. The super-network can be scaled down at various increments, thus creating a series of smaller sub-networks that can be used for inference. Moreover, we employ universally slimmable networks \cite{b22} that dynamically and continuously scale down the number of activation nodes in each hidden layer. The design of techniques to select which sub-networks to use during inference is an area of research attracting considerable interest. The very few available solutions (targeting image classification only), \cite{b21, jiang2023dynamic}, use neural gates to intelligently select sub-networks. With NaviSlim, we advance the state-of-the-art by designing and training a dynamic slimmable neural network for navigation whose shape is controlled by a context-aware gate capable of selecting from a continuous array of sub-networks on a sample-by-sample basis. 

\section{Overview and Problem Formulation}
\label{sec:overview}

Let us first describe the general setting in which we position our contribution. We consider the task to navigate a microdrone with extremely limited onboard resources, in terms of sensing and computing capabilities, from one location to another while avoiding collisions and minimizing path length. Although the framework and methodology we propose are applicable to more general settings, here we consider a micro-drone equipped with: (\emph{a}) multiple depth sensors (\emph{e.g.}, LiDARs) pointing in different directions (\emph{e.g.}, forward and downward), and (\emph{b}) a GPS module returning its position on the map. Depth sensors provide rich information about the environment, which can be used for navigation in settings that do not require high-level reasoning (e.g., semantic features of objects such as the meaning of a traffic sign). Sensor information is input into a neural network which outputs motion commands for the drone.

\begin{figure}[htbp]
\centerline{\includegraphics[width=0.48\textwidth]{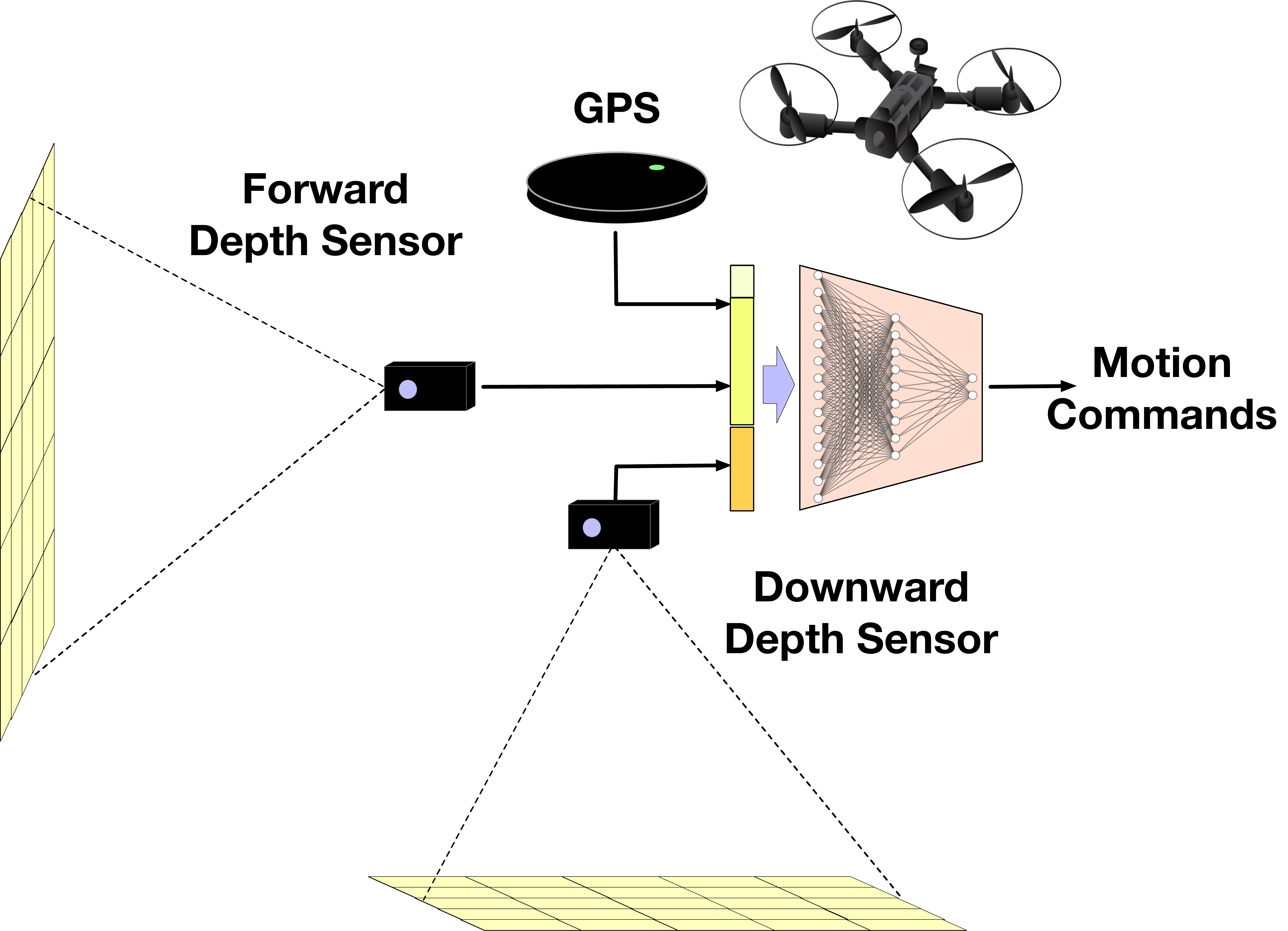}}
\caption{High-level schematics of the considered sensing-computing-control pipeline. The neural network processes the input of forward and downward facing depth sensors and GPS to produce motion commands that fly the drone on an ideally length-optimal path from point A to B.}
\label{schematics}
\end{figure} 

In the context of micro-drones, the energy associated with sensing and computing represents a non-negligible portion of the overall expenditure. Our measurements show that the continuous execution of a relatively lightweight convolutional model use for object detection, executed on a GPU, can take up to 12\% of the total power needed for airborne motion, sensing, and computing. Thus, we consider both dynamic sensors and dynamic neural networks that can be controlled to minimize resource usage. The depth sensors can be tuned to scan a partial field of view, where the smaller the acquired field of view the smaller the time and energy used by the sensor. The relationship between scanned area, resolution, sampling time, and energy is exemplified by lightweight LiDAR sensors that require an amount of time and energy proportional to the extent of the variable scanned area and sampling time \cite{lee2021efficient}. In the framework we propose, the neural network used to control motion commands is a universally slimmable network \cite{b22}, where the number of nodes in each hidden layer can vary to accordingly reduce time and energy spent during computation.

\subsection{Problem Formulation}
\label{subsec:problem}

We consider a drone with an on-board processor tasked to navigate an unknown terrain by utilizing a heterogeneous sensor array and embedded neural model. Given a set of sensor observations, \textbf{o}(t), measured at the current timestep, \textit{t}, the overall objectives of the embedded neural model are:

\vspace{1mm}
\noindent
$\bullet$ {\bf Navigation:} Output navigation motions, \textbf{n}, required for the controller to drive the drone on a length-optimal path that minimizes flight time and energy, while avoiding collisions.

\vspace{1mm}
\noindent
$\bullet$ {\bf Computing:} Execute operations, \textbf{c}, that minimize computing resources used to calculate \textbf{n}. We use the number of active sub-network parameters, \textit{m}, as a proxy for computing resource usage -- an intuitive rationale that we validate experimentally. 

\vspace{1mm}
\noindent
$\bullet$ {\bf Sensing:} Query commands, \textbf{s}, that minimize sensing resources used to acquire observations which are then used as inputs to calculate \textbf{n}. We allocate to the sensors a discrete power level, \textit{w}, used as a proxy for sensing resource usage.

\begin{figure}[htbp]
\centerline{\includegraphics[width=0.48\textwidth]{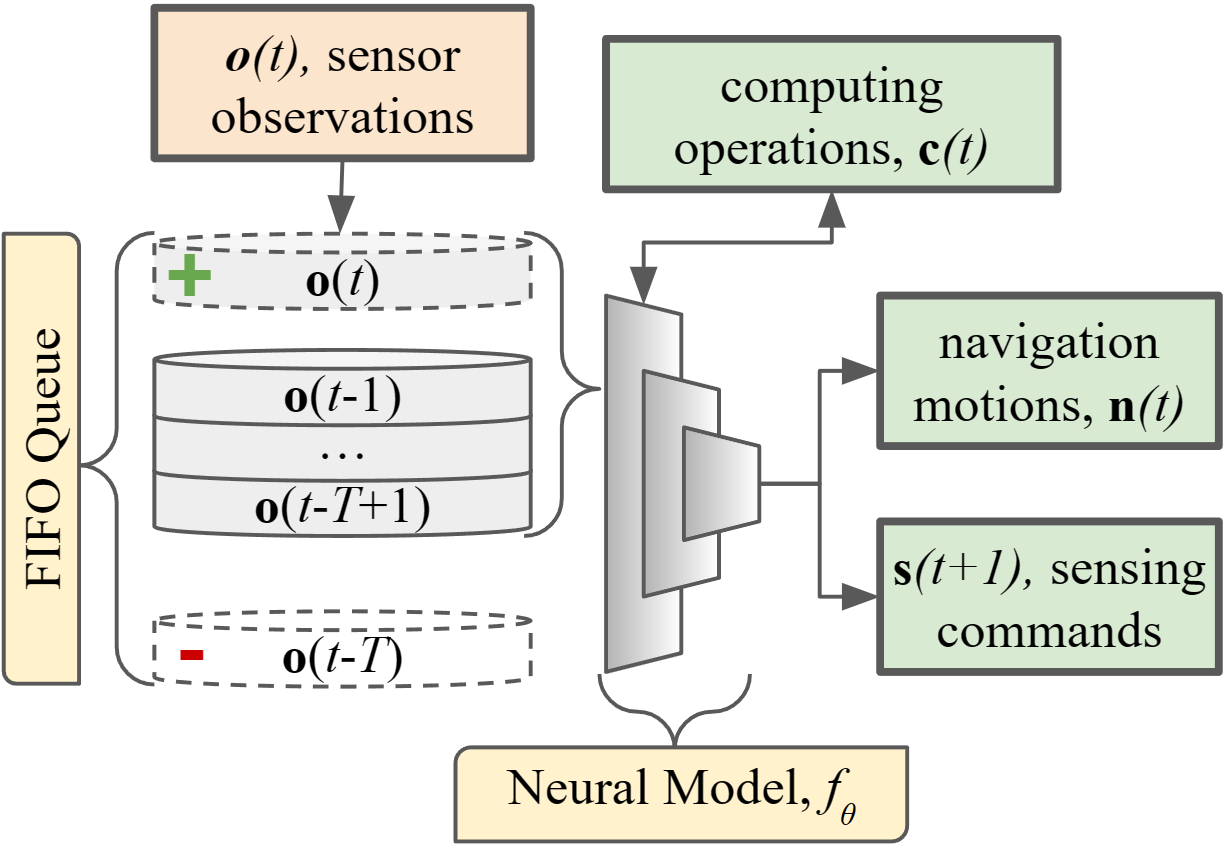}}
\caption{High-level schematics of the considered sensing-computing-control pipeline, now with an introduced attention mechanism and the ability to predict context-aware computing operations and sensing commands.}
\label{genstruc}
\end{figure}

\vspace{1mm}
The pipeline from $\mathbf{o}(t)$ to $[\mathbf{n}(t), \mathbf{c}(t), \mathbf{s}(t+1)]$ is illustrated in
Fig.~\ref{genstruc}. First, the newly acquired set of sensor observations for the current time step, $\mathbf{o}(t)$, is sent into a First In First Out (FIFO) queue. The FIFO queue acts as an attention mechanism, keeping the top \textit{T}-many recent observations as done in \cite{b26} for autonomous control of Atari games. The FIFO queue is then input into a context-aware neural model that: 1) uses an intermediate mechanism to predict the minimum $\mathbf{c}(t)$ needed to predict  $\mathbf{n}(t)$ given the scenario at the current time step, 2) outputs the predicted $\mathbf{n}(t)$ values to execute at the current time step, and 3) outputs the predicted $\mathbf{s}(t+1)$ values to use during sensor acquisition at the next time step.

Here we formalize the problem that we aim to solve. Let $\theta$ be a set of trainable model parameters, such that $\mathbf{a} = f_{\theta}(FIFO)$, where $f_{\theta}$ is the model and $\mathbf{a}$ is some subset of $[\mathbf{n}, \mathbf{c}, \mathbf{s}]$. Let $\mathbf{p}$ be a path taken by the drone where the length of $\mathbf{p}$ is equal to the number of time steps, $P$ be a set of known length-optimal paths, and $\hat{P}$ be the set of paths taken using $f_{\theta}$. We optimize $\theta$ by minimizing the expected trade off between resource costs of computation, \textit{m}, and resource costs of sensing, \textit{w}, as controlled by a scalar, $0 \leq \alpha \leq 1$, and under the constraint that the length of each path taken by $f_\theta$ is no longer than that of that of a scalar, $\beta \geq 1$, times the length of each corresponding optimal path:
\begin{equation}
\label{eq:opt1}
\begin{aligned}
\min_{\theta} \; < \alpha m + (1-\alpha) w >, \\
\textrm{s.t.} \; length(\hat{P}^{(i)}) <= \beta * length(P^{(i)}) \; \forall \; i \in \{1, ..., b\},
\end{aligned}
\end{equation}
where $<>$ indicates the expected value over all time steps and paths, $b$ is the total number of paths, $P^{(i)}$ indicates the $i^{th}$ path, and $length(\mathbf{p})$ indicates the number of time steps in path $\mathbf{p}$. We design Equation~(\ref{eq:opt1}) as to minimize computing and sensing resource usage, while retaining navigation accuracy. The constraint in Equation~(\ref{eq:opt1}) is required, otherwise the optimization problem would result in the trivial solution where computing and sensing parameters are equal to zero.

\subsection{Challenges and Contributions}
\label{sec:challenges}

The two main challenges in developing \emph{NaviSlim} are:

\vspace{1mm}
\noindent
{\bf Test Bed Environment:} To train an adaptive model for microdrone navigation, we need a test bed environment to allow the algorithm to accumulate a large amount of experience across settings with a broad range of complexity and maneuvers. To this end, we developed our open-source software module to interface our models with Microsoft AirSim \cite{b3}, which is a robust drone simulator for graphics rendering, sensor acquisition, and physics handling. Our interface obfuscates the environment to the model during training and at deployment time, so that \emph{NaviSlim} can be seamlessly ported to various simulation and real-world environments.

\vspace{1mm}
\noindent
{\bf Model Design and Training Procedure:} A fundamental question is how to design and train a neural model that can accomplish all three objectives of controlling navigation, computing, and sensing. We base our model design on dynamic slimmable neural networks, and add capabilities to adapt computing and sensing on a sample-by-sample basis. As expected, a single neural network trained from scratch can not converge to any meaningful results that accomplishes all three of these objectives simultaneously. Our solution is to decouple each of these objectives into three respective modules. Each module has with it unique challenges, for which we develop solutions utilizing several methods such as: shortest path algorithms, supervised learning, deep reinforcement learning, curriculum learning, and of most significance knowledge distillation \cite{b31}.

\vspace{2mm}
To the best of our knowledge, the one presented herein is the first neural model that fuses universally slimmable networks with dynamic neural networks to accomplish navigational goals under severe resource constraints, while exploring dynamic sensor scaling, a widely open area of investigation. The core contribution of this paper is \emph{NaviSlim}: a novel framework to design and train neural models that can seamlessly and dynamically adapt their characteristics to environmental context and mission progress to parsimoniously use computation and sensing resources while maintaining high navigation accuracy.

\section{Test Bed Environment}
\label{sec:environment}

We train and test our models with a simulation framework that utilizes Microsoft AirSim \cite{b3}, a robust drone simulator that renders physics and graphics in Unreal Engine \cite{unrealengine}. AirSim has a Application Programming Interface (API) for Python, that can be used to communicate with the simulation, such as: create sensors and acquire observations, issue drone commands, and detect collisions. We use the AirSim API to interface with our \emph{NaviSlim} repository, also in Python, which includes methods for deep reinforcement learning that partially utilize the Stable-Baselines3 library \cite{b24}, neural network implementations that partially utilize the PyTorch library \cite{paszke2017automatic}, and others such as curriculum learning, shortest path algorithms, supervised learning, knowledge distillation, logging, customization, and deployment to other environments  including real world drone controllers. Previous studies have shown capabilities of launching models trained in simulation into the real world \cite{b35, b36} -- thus a simulation tool is a robust means to explore and develop novel model architectures. Using a simulation also mitigates difficulties in training a model with real world hardware that would require mechanisms for episodic deep reinforcement learning.

\begin{figure}[htbp]
\centerline{\includegraphics[width=0.5\textwidth]{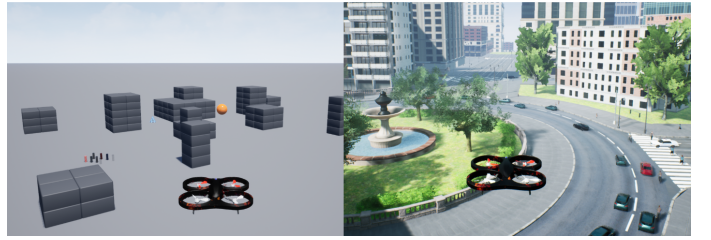}}
\caption{Two maps from Microsoft AirSim: on the left is "Blocks" which contains static objects with arbitrary shapes and sizes, and on the right is "City" which contains both static and dynamic objects expected to be encountered in the real world.}
\label{airsim2}
\end{figure}

Fig.~\ref{airsim2} shows two maps used by AirSim. The first map is "Blocks", which contains several static objects that have generic shapes and sizes. The second map is "City", which contains various static and dynamic objects that have specific shapes and sizes which reflect real world encounters such as buildings, signs, cars, people, and live traffic. Both maps have varying densities of objects, and we train and evaluate over a wide range of these densities.

\section{NaviSlim: Design Overview}
\label{sec:impementation}
In this section, we provide an overview of \emph{NaviSlim}, and will detail the specific components (the navigation and auxiliary models) in the next sections. A key novelty is that we design an auxiliary module to control resource expenditure (\textbf{c} or \textbf{s}), while the navigation module is used to control drone motions (\textbf{n}). Thus \textit{NaviSlim}, $f_\theta$, now consists of the navigation model, $g_\phi$, and the auxiliary model, $h_\psi$. 
If a vanilla approach is taken to train the overall model to simultaneously control both sensing (input) and computing (intermediate calculations used by the model), then the learning process is highly unstable and does not converge to a meaningful control logic -- \textit{i.e.}, the navigation paths fail when evaluated in the test bed environment. Thus, our solution is to decouple computing and sensing into two variants of \emph{NaviSlim}. We refer to methods and models related to computing as \emph{NaviSlim-C}, and those related to sensing as \emph{NaviSlim-S} (see Equation~(\ref{eq:finalmodel})):
\begin{equation}
\label{eq:finalmodel}
\begin{aligned}
\text{\emph{NaviSlim-C}: } \; \; \; \mathbf{n} = g_\phi(FIFO, \; \mathbf{c}=h_\psi(FIFO)) \\
\text{\emph{NaviSlim-S}: } \; \; \; [\mathbf{n}, \mathbf{s}] = [g_\phi(FIFO), h_\psi(FIFO)].
\end{aligned}
\end{equation}
 Note that this structure requires $g_\phi$ and $h_\psi$ to be executed in series for \emph{NaviSlim-C}, while they can be executed in parallel for \emph{NaviSlim-S}. 
 
\begin{figure}[htbp]
\centerline{\includegraphics[width=0.48\textwidth]{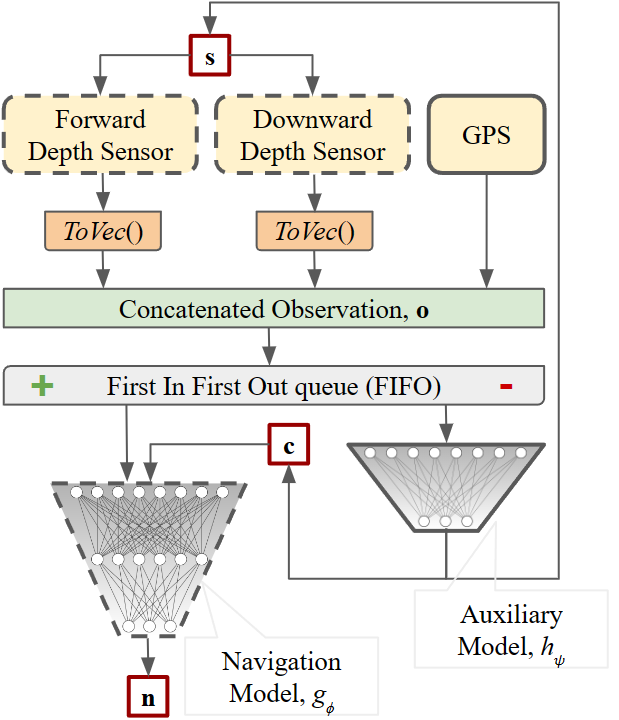}}
\caption{\emph{NaviSlim}: our novel solution for a context-aware framework capable of adapting resource allocation to that which is required by the difficulty of the current scenario. Shown is our specific implementation. The shapes with dotted lines represent components capable of adaptable resource allocation.}
\label{sys}
\end{figure}

The overall system \emph{NaviSlim} model (composed of the navigation and auxiliary models) is illustrated in Fig.~\ref{sys}. The "ToVec()" component converts the data acquired from each depth sensor into preliminary vectors that are then concatenated with the GPS data into one feature vector, \textbf{o}, as measured at time \textit{t}. This concatenated feature vector is then inserted onto the FIFO queue as illustrated previously in Fig.~\ref{genstruc}.

Algorithm~\ref{alg:root} shows how a path, also called an episode, is executed using \textit{NaviSlim}. Included in Algorithm~\ref{alg:root} are various variables used for deep reinforcement learning, as detailed later in Section~\ref{sec:auxiliary}. 

\begin{algorithm}
\caption{Executing an Episode/Path with \textit{NaviSlim}} \label{alg:root}
\textbf{Input:} spawn and goal, max time steps $\tau$, goal tolerance $\eta$, navigation model $g_\phi$, auxiliary model $h_\psi$, reward function.
\begin{algorithmic}
\State set \textbf{c} and \textbf{s} to that which corresponds to maximum resources
\State $t = 1; \; continue=True$
\While{$continue$}
\State acquire sensor observations, $\mathbf{o}$, given \textbf{s}
\State add $\mathbf{o}$ to $FIFO$
\If{\emph{NaviSlim-C}}
\State $\mathbf{a} \coloneqq \mathbf{c} = h_\psi(FIFO)$
\EndIf
\If{\emph{NaviSlim-S}}
\State $\mathbf{a} \coloneqq \mathbf{s} = h_\psi(FIFO)$
\EndIf
\State $\mathbf{n} = g_\phi(FIFO)$, given \textbf{c}
\State move drone using \textbf{n}
\State calculate reward, $r$
\State t = t + 1
\If{collision detected} 
    \State $termination = collision; \; continue = False$
\EndIf
\If{distance to target position $ < \eta$} 
    \State $termination = goal; \; continue = False$
\EndIf
\If{$t > \tau$} 
    \State $termination = time; \; continue = False$
\EndIf
\EndWhile
\end{algorithmic}
\textbf{Return:} $E = [[\mathbf{o}, \mathbf{n}, \mathbf{a}, r]^{(j)} \; \forall \; j \in \{1, ..., t-1\}, termination]$
\end{algorithm}

\subsection{Universally Slimmable Networks}
\label{sec:universally}

The navigation model is a universally slimmable network \cite{b22}. Here we introduce a new variable called the slimming factor, $\rho$, which controls the number of active nodes in each hidden layer, that is, $\mathbf{c} = [\rho]$ since $\rho$ controls the number of operations required to execute the navigation model.

Take an arbitrary hidden layer comprised of a vector of nodes, \textbf{h}, indexed from the node at position $k=1$ to $k=q$ where $q$ is the maximum number of nodes available for that hidden layer. The quantity $q$ corresponds to the number of hidden layer nodes used by the static "super-network", which in our context is a network whose number of parameters must match the most difficult operating scenario. This super-network persists when $\rho=1$, whereas a sub-network is activated when $\rho<1$.

\begin{figure}[htbp]
\centerline{\includegraphics[width=0.5\textwidth]{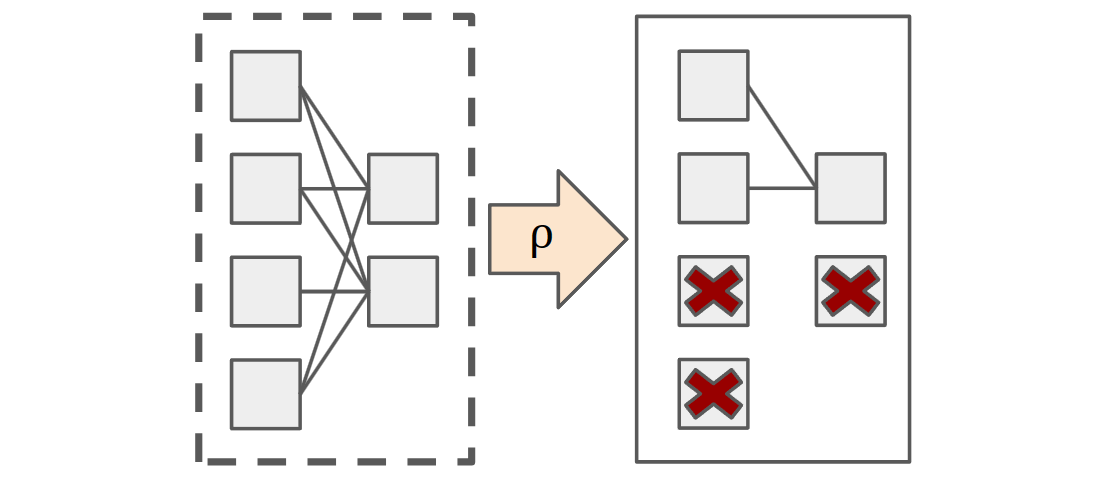}}
\caption{Procedure used to "slim" a universally slimmable neural network. The variable $\rho$ is used to scale the number of active nodes in each hidden layer. In this example, there are 4 nodes in the first hidden layer and 2 in the second -- this is the super-network. Also in this example, we use $\rho=0.3$ so that 2 nodes are deactivated in the first layer and 1 in the second -- this is a sub-network. Note that all weights connected to the deactivated nodes, represented by the lines in between the hidden layers, are also severed.}
\label{slimnet}
\end{figure}

The number of active nodes in a layer is equal to $roof(\rho q)$, where $roof()$ rounds up to the nearest integer value, such that the active nodes are those indexed in the range \textbf{[}$1, roof(\rho q)$\textbf{)} and the deactivated nodes are those indexed in the range \textbf{[}$roof(\rho q), q$\textbf{]}. Such a procedural deactivation, as opposed to a random Bernoulli distribution such as used in dropout, is required so that we can select specific sub-networks from the super-network for both training and inference (see Fig.~\ref{slimnet}).

The relationship between $\rho$ and the total number of active parameters, \textit{m}, in a sub-network is quadratic. Let \textit{u} bet the number of input layer nodes, \textit{l} be the number of hidden layers, \textbf{q} be an \textit{l}-length vector containing the number of nodes in each hidden layer, and \textit{v} be the number of output nodes. We consider a fully connected feed-forward multi-layer perceptron with bias terms and at least one hidden layer. We can directly calculate the number of active parameters in a sub-network, \textit{m}, as a function of $\rho$:
\begin{equation}
\label{eq:rho}
\begin{aligned}
m(\rho) = u \rho q_1 + \rho q_1 + \rho q_1 \rho q_2 + \rho q_2 + ... + \rho q_l v + v, \\
m(\rho) = (\Sigma_{i=1}^{l-1} q_i q_{i+1}) \rho^2 + (u q_1 + v q_l + \Sigma_{i=1}^l q_i) \rho + v.
\end{aligned}
\end{equation}
Thus, there is a quadratic decrease in the number of active parameters of a sub-network with a decreased $\rho$.

\subsection{Supervised Learning with Knowledge Distillation}
\label{sec:distillation}

A key method used when training \emph{NaviSlim} is knowledge distillation. From the perspective of a slimmable network, the main idea of knowledge distillation is to teach sub-networks similar outputs as the super-network. This is accomplished by adding a step to typical supervised learning after the error gradient is calculated between ground truth labels ("hard targets") and the outputs of the super-network , that calculates an additional error gradient between the outputs of the super-network ("soft targets") and outputs of any activated sub-networks. The loss function uses the error between both: A) super-network outputs and ground truth labels and B) sub-network outputs and super-network outputs. This combined supervised learning and knowledge distillation step, what we refer to as the $supervised\_distillation$ function, is defined differently between \emph{NaviSlim-C} and \emph{NaviSlim-S}.

\subsection{NaviSlim-C}
\label{sec:computing}

The objective of \emph{NaviSlim-C} is to reduce  computing resources spent during execution of the navigation model. This is accomplished by defining the $supervised\_distillation$ function with Algorithm~\ref{alg:computation_navigation}. Algorithm~\ref{alg:computation_navigation}
uses the "sandwich rule" \cite{b22}, a method for sampling $\rho$ when training with knowledge distillation by calculating the error gradient first with $\rho=1$, distilling with $\rho$ set to the minimal value, and then distilling with some random intermediate values of $\rho$.

\begin{algorithm}
\caption{$supervised\_distillation$ for \emph{NaviSlim-C}} \label{alg:computation_navigation}
\textbf{Input:} navigation model in training $g_\phi$, input $FIFO$, target output $\mathbf{n}$, \textit{loss} function.
\begin{algorithmic}
\State set running error gradient, \textit{grad}, to zero
\State $\hat{\mathbf{n}} = g_\phi(FIFO, \rho=1)$ \Comment{Max Resource Cost}
\State aggregate \textit{grad} with \textit{loss} given  $\mathbf{n}$ and $\hat{\mathbf{n}}$ \Comment{Supervised}
\State $\hat{\mathbf{n}}' = g_\phi(FIFO, \rho=0.125)$ \Comment{Min Resource Cost}
\State aggregate \textit{grad} with \textit{loss} given $\hat{\mathbf{n}}$ and $\hat{\mathbf{n}}'$ \Comment{Distillation}
\For{a number of times, \textit{i.e.} 2}
\State $\hat{\mathbf{n}}' = g_\phi(FIFO, \rho \sim U(0.125,1))$ \Comment{Random Cost}
\State aggregate \textit{grad} with \textit{loss} given $\hat{\mathbf{n}}$ and $\hat{\mathbf{n}}'$  \Comment{Distillation}
\EndFor
\end{algorithmic}
\textbf{Return:} \textit{grad} 
\end{algorithm}

\subsection{NaviSlim-S}
\label{sec:sensing}

The objective of \emph{NaviSlim-S} is to minimize the resource usage of sensor acquisition at the next time step. This is accomplished by defining the $supervised\_distillation$ method with Algorithm~\ref{alg:sensing_navigation}: a novel approach applied to sensing, that distills the network to have similar output regardless of the variable sensor array power level, \textit{w}. 

We introduce two variables used to control the respective power levels of the forward, $p_f$, and downward, $p_d$, facing depth sensors, with $\mathbf{s} = [p_f, p_d]$. The range of $p_f$ is $[1, 3]$, where the value one corresponds to the minimal power level which uses the smallest area available for scanning. The range of $p_d$ is $[0, 3]$, where zero is the minimal power level corresponding to completely turning off the downward depth sensor. Increasing power levels corresponds to the acquisition of larger areas during scanning with the respective sensor, where a power level of three is the maximum area, and thus the maximum power and time expenditure. Note that $p_f >= 1$ so at-least some sensor information can be acquired at all times. The input layer to the navigation network is designed so that the number of required input nodes corresponds to the magnitude of $p_f$ and $p_d$. Input nodes are deactivated using a procedural approach, so that sub-networks consisting of subsets of input nodes can be selected during inference and training, similar to \emph{NaviSlim-C}.

\begin{algorithm}
\caption{$supervised\_distillation$ for \emph{NaviSlim-S}} \label{alg:sensing_navigation}
\textbf{Input:} navigation model in training $g_\phi$, input $FIFO$ (at max power level), target output $\mathbf{n}$, \textit{loss} function.
\begin{algorithmic}
\State set running error gradient, \textit{grad}, to zero
\State $\hat{\mathbf{n}} = g_\phi(FIFO)$ \Comment{Max Resource Cost}
\State aggregate \textit{grad} with \textit{loss} given  $\mathbf{n}$ and $\hat{\mathbf{n}}$ \Comment{Supervised}
\State $FIFO'$ = re-sampled all \textbf{o} in $FIFO$ with $p_f=1, p_d=0$
\State $\hat{\mathbf{n}}' = g_\phi(FIFO')$ \Comment{Min Resource Cost}
\State aggregate \textit{grad} with \textit{loss} given $\hat{\mathbf{n}}$ and $\hat{\mathbf{n}}'$ \Comment{Distillation}
\For{a number of times, \textit{i.e.} 2}
\State $FIFO'$ = re-sampled with \textit{T}-many heterogeneous \State \hspace{\algorithmicindent} random $p_f$ and $p_d$ values applied to each \textbf{o} in \State \hspace{\algorithmicindent} $FIFO$, where $p_f \sim U(1,3)$ and $p_d \sim U(0,3)$
\State $\hat{\mathbf{n}}' = g_\phi(FIFO')$ \Comment{Random Resource Costs}
\State aggregate \textit{grad} with \textit{loss} given $\hat{\mathbf{n}}$ and $\hat{\mathbf{n}}'$  \Comment{Distillation}
\EndFor
\end{algorithmic}
\textbf{Return:} \textit{grad} 
\end{algorithm}

Algorithm~\ref{alg:computation_navigation}
and Algorithm~\ref{alg:sensing_navigation} share a similar structure based on knowledge distillation and the sandwich rule. The difference between the two algorithms is how the respective networks are distilled. Algorithm~\ref{alg:computation_navigation} distills the navigation network to operate with a varying number of nodes in each hidden layer, while Algorithm~\ref{alg:sensing_navigation} distills the network to operate with varying power levels used to acquire input sensor observations (which is likened to slimming the input layer). We extrapolate that similar approaches can be used to distill the network to operate with other varying attributes. This is why we use the name \emph{NaviSlim} for both the variants, as \emph{NaviSlim} fuses navigation with a form of "slimming" applied to different attributes of the navigation network.

\section{NaviSlim: Navigation Module}
\label{sec:navigation}

The objectives of the navigation module are two fold: 1) navigate along a length-optimal path from a spawn position to the goal, while avoiding collisions, and 2) execute the underlying navigation model using variable sensing and computing parameters. The process for training the underlying neural network is outlined below:

\begin{enumerate}
    \item Collect ground truth length-optimal paths, $P$, using an A-star \cite{b27} shortest path algorithm.
    \item Acquire sensor observations at each time step in $P$.
    \item Use supervised learning with knowledge distillation to train a dynamic neural network that maps the FIFO queue of recent observations to navigation motions, \textbf{n}.
    \item Freeze the navigation module, as to no longer update the trainable network parameters.
    \item Evaluate for a successful model by deploying $g_\phi$ to the test bed environment.
\end{enumerate}

\subsection{A-star Shortest Path Algorithm}
\label{sec:astar}
Shortest path algorithms are used to solve the problem of finding length-optimal paths between two points. Graph algorithms are a subset of shortest path algorithms where the problem can be constructed in a graph structure with vertices and edges. A-star \cite{b27} is a flavor of graph shortest path algorithms, that is guaranteed to find the optimal solution without having to traverse every possible path. We implement A-star by reconstructing the Blocks map into a graph where each vertex corresponds to a spatial point on the map, and edges define if an adjacent position is valid (\textit{i.e.}, does not have an object in it). The cost of a path is simply the total distance traveled. 

We partition the Blocks map into regions used for a training, validation, and testing set. We then use A-star to find the optimal paths, \textit{P}, in each set. The test bed environment is used to collect simulated sensor observations from Microsoft AirSim at each timestep in \textit{P}. This results in a dataset of hard targets that map $FIFO \mapsto \mathbf{n}$ for each time step in each path within \textit{P}. These are then used to train the navigation network.

\subsection{Training the Neural Network}
\label{sec:training}

Given a set of maps, $FIFO \mapsto \mathbf{n}$, over optimal paths, \textit{P}, the training procedure for the navigation neural network, $g_\phi$, is outlined in Algorithm~\ref{alg:nav}. The function $supervised\_distillation$ is an argument to Algorithm~\ref{alg:nav} that controls how supervised learning works in unison with knowledge distillation as earlier defined in Section~\ref{sec:distillation}. The training set is used to drive the error gradient, the validation set is used to trigger early stopping to mitigate overfit, and the test set is held out to later evaluate the navigation model. We use mean squared error as our loss function.

\begin{algorithm}
\caption{Training a Navigation Neural Network} \label{alg:nav}
\textbf{Input:} training set, validation set, randomly initialized $g_\phi$, $supervised\_distillation$ function, $loss$ function.
\begin{algorithmic}
\While{not converged}
\State set running error gradient to zero
\State sample batch of [$FIFO \mapsto \mathbf{n}$] pairs from training set
\For {each $FIFO \mapsto \mathbf{n}$ in batch...}
\State grad = $supervised\_distillation(g_\phi, FIFO, \mathbf{n}, loss)$
\State aggregate grad into running error gradient
\EndFor
\State update $\phi$ using optimizer with aggregated error gradient
\State use the validation set to check for early stopping
\EndWhile
\end{algorithmic}
\textbf{Return:} trained $g_\phi$
\end{algorithm}

After the navigation neural network is trained using Algorithm~\ref{alg:nav}, the optimized parameters, $\phi$, are frozen and not updated again. We evaluate if a trained navigation model is successful by deploying $g_\phi$ into the AirSim environment and measuring the percent of paths in the test set that successfully reach their goal -- when using the super-network and Algorithm~\ref{alg:root} but without an auxiliary module, $h_\psi$. Thus, we are only evaluating the navigation prowess of the static super-network without adaptation. 

\section{NaviSlim: Auxiliary Module}
\label{sec:auxiliary}

The objective of the auxiliary module is to control adaptation of either \textbf{c} or \textbf{s} as applied to the navigation module. The following steps are taken to create the auxiliary module:

\begin{enumerate}
    \item Successfully train a navigation model, $g_\phi$.
    \item Create a reward function that penalizes the navigation algorithm for taking sub-optimal paths. 
    \item Train the auxiliary model, $h_\psi$, using a Twin Delayed Deep Deterministic Policy Gradient (TD3) \cite{b11} deep reinforcement learning algorithm. The objective is to map the FIFO queue to adaptation parameters, \textbf{c} or \textbf{s}.
    \item Evaluate for a successful model by deploying $h_\psi$ with $g_\phi$ to the test bed environment.
\end{enumerate}

\subsection{Deep Reinforcement Learning}
\label{sec:reinforcement}

Deep Reinforcement Learning (DRL) consists of an agent that will be episodically traversing an environment during the training process, by taking actions which result in an evaluated reward. The policy is learned during training that maps observations to optimal actions which in essence maximize the rewards. The word "deep" simply refers to using a deep neural network as the policy mechanism. 

\subsection{Q-learning}
\label{sec:qlearning}

The Q-value, derived from the Bellman equation \cite{tesauro1995temporal}, is an estimation of the aggregation of immediate and long term rewards in an episode -- where higher Q-values correspond to more effective actions. Given a reward function, $r(state)$, that inputs the \textit{state} found after executing \textbf{n}, Equation~(\ref{eq:opt2}) shows the optimization problem used to train  $h_\psi$:

\begin{equation}
\label{eq:opt2}
\begin{aligned}
\max_{\psi} \; <\text{\textit{Q-value}}>, \\
\text{\textit{Q-value}}(\mathbf{p}, t) =  \Sigma_{i=t}^{length(\mathbf{p})} \gamma^{i-t} * r(state^{(i)}),
\end{aligned}
\end{equation}
where $<>$ denotes the expected Q-value resulting from using $g_\psi$ over all paths and time steps, $state^{(i)}$ corresponds to state variables used to calculate the reward at the $i^{th}$ time step from path $\mathbf{p}$, and $\gamma$ is a parameter called the discount factor which applies a decaying penalty to long term rewards.

\subsection{Reward Function}
\label{sec:reward}

Designing a reward function is highly non-trivial, as it directly affects the stability and convergence of the training algorithm, along with the learned policy. We use a random walk algorithm to estimate the behavior of a proposed reward function. Each iteration takes a mean step closer to an arbitrary goal initially set 100 meters away, with Gaussian noise added at each step. We then recursively calculate the rewards at each step to estimate the Q-value corresponding to the theoretical episode. This process is used to design and adjust the coefficients in the reward function by evaluating them against the average of several random walks. We, then, select the following reward function:
\begin{equation}
\label{eq:rew}
r {=} \left\{
\begin{array}{ll}
      - \lambda_o & collision \\
      \lambda_g & goal \\
      \lambda_d tanh(d) {-} \lambda_t {-} \lambda_c \rho  {-} \lambda_s (p_f  {+}  p_d) & otherwise
\end{array},
\right. 
\end{equation}
where the constants $\lambda_x$ are non-negative weights applied to different state variables. The first two conditions assign large reward values to terminal states that are encountered either when the drone reaches the goal or collides with an object. The third condition assigns a set of penalties applied to intermediate states to encourage shorter paths. The first term of the third condition applies either a reward or penalty based on $d$, the change in distance between the drone and goal in between time steps. The second term is a constant penalty for the number of steps taken in an episode. The third term applies a penalty based on the amount of resources allocated to sensing, as controlled by $\rho$. The fourth term  applies a penalty based on the amount of resources allocated to computing, as controlled by $p_f$ and $p_d$.

\subsection{TD3}
\label{sec:td3}

Among the many available DRL algorithms, we select a Twin Delayed Deep Deterministic Policy Gradient (TD3) \cite{b11} algorithm; as it allows a continuous multivariate observation space, a continuous multivariate action space, and outperforms other DRL algorithms that we tested. TD3 is based on an actor-critic double deep Q-learning paradigm with clipped action noise and delayed policy updates, consisting of a series of six neural networks used in the training process. The actor network, $h_\psi$, is the policy mechanism that inputs the FIFO queue of recent observations and outputs either \textbf{c} or \textbf{s}. The target actor network is used to estimate the next action, while noise is added to that action to smooth out training and account for error. The critic network estimates the Q-value, since future states are unknown. The target critic network is used to measure error in the critic network, since we can not calculate all possible future states. The target critic and target actor networks are initially clones of the critic and actor network, respectively, but the weights vary over time and every few episodes the target network weights are updated using a Polyak weighted average. 

Two critic networks are used to help stabilize training by selecting the minimum Q-value between both critics, which is referred to as double deep Q-learning \cite{van2016deep}. When training a neural network with a TD3 algorithm, initially actions are taken at random to explore the environment. Then the neural network is increasingly used to predict actions as to exploit the policy being learned. Even though six neural networks are used in the training process, only the actor network, $h_\psi$, is executed at the deployment stage.

\subsection{Curriculum Learning}
\label{sec:curriculum}

A common difficulty in training a neural network with DRL arises when the initial task is too difficult. This causes poor rewards early in the learning process which can stagnate training. A solution is to start with an easier task, then progressively increase the difficulty. We implement curriculum learning by starting with a small distance between spawn and goal positions, then incrementally increase this distance as enough evaluation paths successfully reach the goal.

\subsection{Training the Neural Network}
\label{sec:training2}

We train the auxiliary network, $h_\psi$, using Algorithm~\ref{alg:td3}. Training the auxiliary network is the main bottleneck for training \emph{NaviSlim}, because it can take several days before finishing. Further, the parameters to Algorithm~\ref{alg:td3} are highly sensitive (some of which are omitted), which requires ample time to explore them. Having several computers to run training in parallel with different parameters is highly advantageous. We evaluate if a trained auxiliary model is successful by deploying $g_\phi$ and $h_\psi$ into the AirSim environment and measuring the percent of paths in the test set that successfully reach their goal when using Algorithm~\ref{alg:root}. Further, we insure the constraint holds in Equation~\ref{eq:opt1} since this is not guaranteed by Equation~\ref{eq:opt2}.

\begin{algorithm}
\caption{Training an Arbitrary Auxiliary Network} \label{alg:td3}
\textbf{Input:} training region, evaluation set, trained navigation network $g_\phi$, randomly initialized auxiliary network $h_\psi$
\begin{algorithmic}
\State create a replay buffer of set length to store episodic data
\State deep copy $h_\psi$ to make target actor network
\If{\emph{NaviSlim-C}}
\State $\mathbf{a} \coloneqq \mathbf{c}$ \Comment{Actions will Reduce Computational Costs}
\EndIf
\If{\emph{NaviSlim-S}}
\State $\mathbf{a} \coloneqq \mathbf{s}$ \Comment{Actions will Reduce Sensing Costs}
\EndIf
\State let $h_{rand}$ be a random generator of \textbf{a}
\State randomly initialize 2 critic networks and clone target critics
\State Each '?'-Boolean bellow is parameterized by $i$.
\For{$i=1$; $i <= $ max episodes; $i=i+1$} 
\State generate random spawns and goals from the train region
\If{explore?}
\State \textit{E} = Algorithm~\ref{alg:root} with $h_{rand}$  \Comment{exploration}
\Else
\State \textit{E} = Algorithm~\ref{alg:root} with $h_\psi$ \Comment{exploitation}
\EndIf
\State add (E, t) data from each time step in \textit{E} to replay buffer
\If{train?}
\State sample batch of $(E, t)$ data from replay buffer
\For{each $(E, t)$ data in batch} 
\State $\tilde{\mathbf{a}} = $ 
target\_actor($LIFO_{t+1}$) 
\State $\tilde{\mathbf{a}} = \tilde{\mathbf{a}}$ + clipped Gaussian noise
\State $q = r_t + \gamma \; *$ min(target\_critics($LIFO_{t+1}$, $\tilde{\mathbf{a}}$))
\For{each critic}
\State $\hat{q} = $ critic($LIFO_t$, $\mathbf{a}_t$)
\State calculate $loss$ between $q$ and $\hat{q}$
\State update critic with optimizer and $loss$
\EndFor
\If{update?}
\State $loss$ = -1 * critics[0]($LIFO_t$, $\mathbf{a}_t$).mean()
\State update $h_\psi$ with optimizer and $loss$
\State Polyak update target actor and target critics
\EndIf
\EndFor
\EndIf
\If{evaluate?}
\For{each spawn and goal pair in evaluation set}
\State \textit{E} = Algorithm~\ref{alg:root} with $h_\psi$
\EndFor  
\If{enough evaluation paths successful}
\State increase distance between spawn and goal
\State terminate training if that distance is large enough
\EndIf
\EndIf
\EndFor
\end{algorithmic}
\textbf{Return:} trained $h_\psi$
\end{algorithm}

\section{Results}
\label{sec:results}

\begin{figure}[htbp]
\centerline{\includegraphics{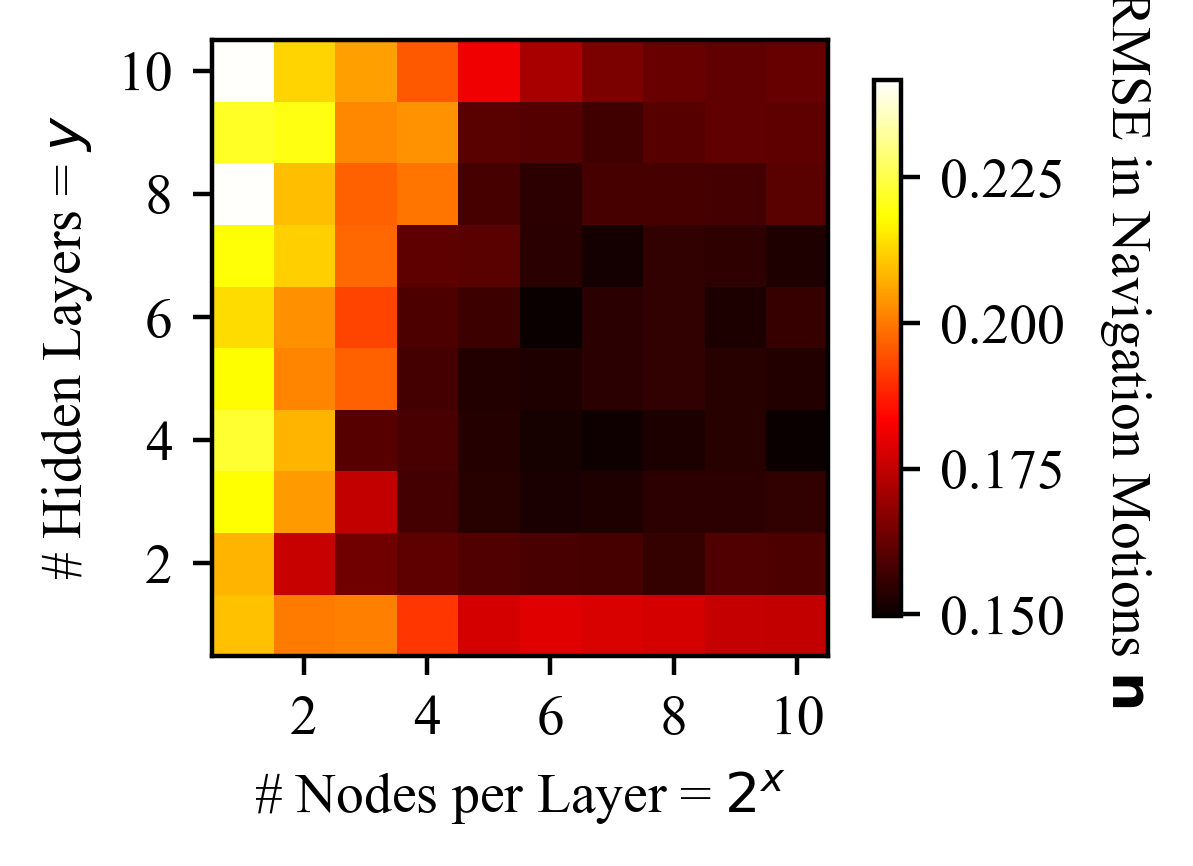}}
\caption{Results of a hyper-parameter grid search that explores different neural network sizes of the navigation model, $g_\phi$. Shown is the Root Mean Squared Error (RMSE) calculated between the length-optimal navigation motions found from A-star and those predicted from a trained $g_\phi$.}
\label{hyper}
\end{figure}

First we explore the size, the number of hidden layers and nodes in each layer, of the navigation super-network. We test each configuration by training the navigation model 10 random times with different seeds, and take the seed with the best error as measured from a static test set. Fig.~\ref{hyper} shows results of a hyper parameter grid search, displaying the Root Mean Squared Error (RMSE) between length-optimal motions as found from A-star, $\mathbf{n}$, and those predicted from $g_\phi$, $\hat{\mathbf{n}}$. The darker region of Fig.~\ref{hyper} indicates that, as expected, using a larger super-network results in a reduced navigation error. The navigation model is then deployed in the simulation test bed, using Microsoft AirSim. 

\begin{figure}[htbp]
\centerline{\includegraphics{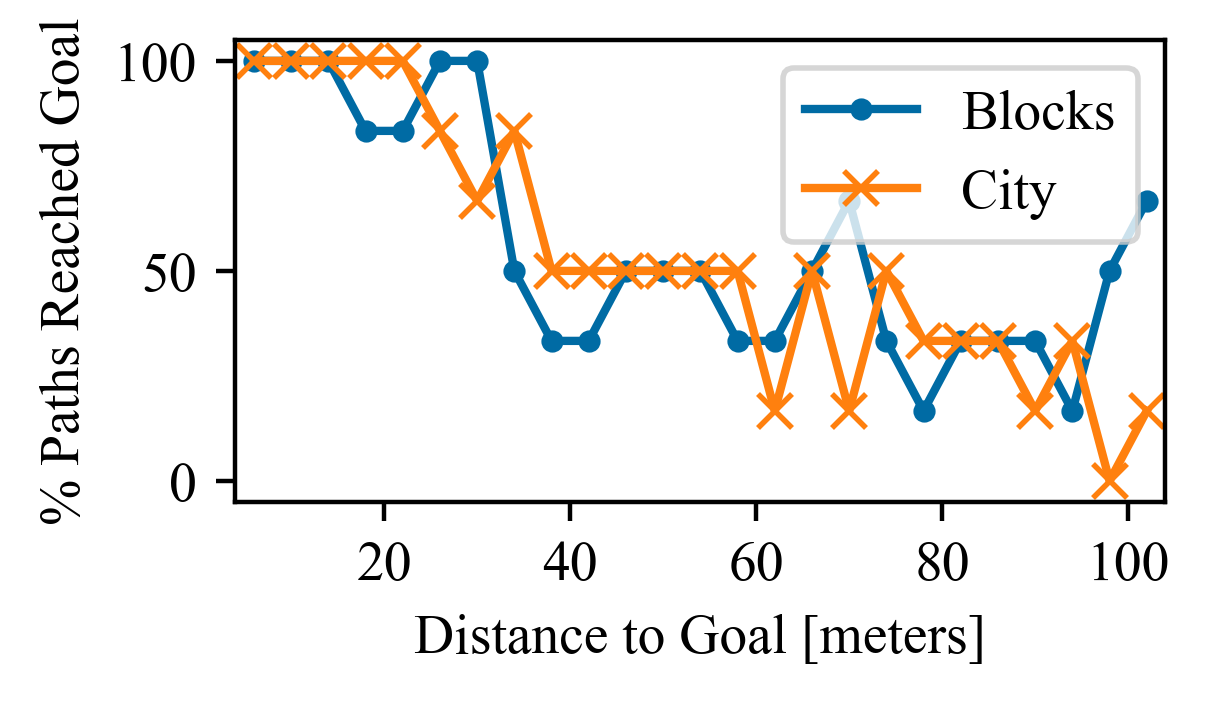}}
\caption{Results of a trained navigation model, $g_\phi$, evaluated in the test bed environment using both AirSim maps. This figure shows the percent of evaluation paths that reach their goal as a function of starting distance.}
\label{pretrain2}
\end{figure}

Fig.~\ref{pretrain2} shows the percentage of evaluation paths which successfully reach the goal versus distance to goal. The percentage of successful paths drops with increasing distance, as expected. The navigation models perform remarkably well when deployed to the more complex City map even though they are only trained using samples from the simple Blocks map. This illustrates the prowess, and generalization, of our navigation training methods. Typical approaches would stop here, and have a static navigation network with fixed computing and sensing required by that of the most difficult scenario. However, the following results come from experimentation of our novel approach to adapt computing and sensing with \emph{NaviSlim}.

\begin{figure*}[htbp]
\centerline{\includegraphics{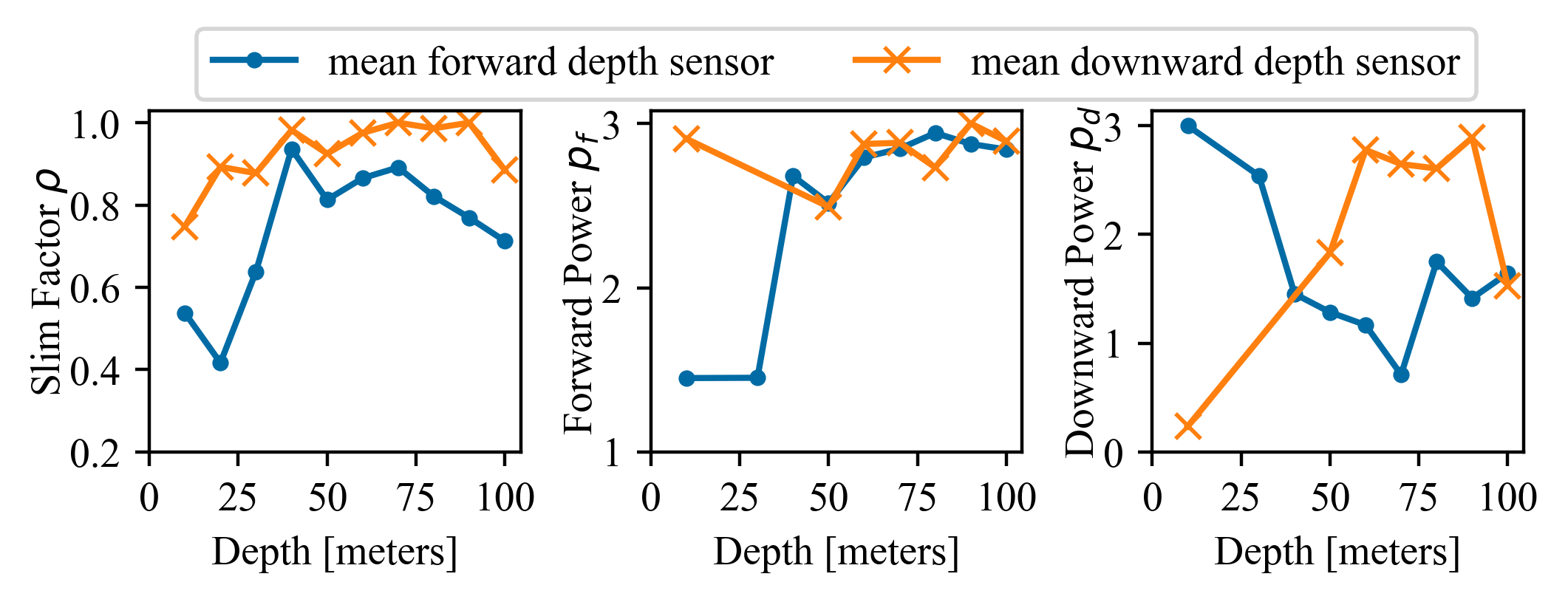}}
\caption{Mean results of the different adaptability variables that control various resource allocations as predicted from the auxiliary network, $h_\psi$, over all values from successful paths in the test set. A larger slim factor, $\rho$, corresponds to increased computation costs (we find this to be time and energy with fixed power) needed to run the navigation network, $g_\phi$. A larger power level, either $p_f$ for the forward facing depth sensor or $p_d$ for the downward facing depth sensor, corresponds to increased sensing costs (proven in literature to be both time and power \cite{lee2021efficient}) needed to acquire observations from the sensor array. The x-axis shows the mean values returned from each depth sensor, in meters, and is binned at increments of 10 meters (note that some bins are missing, this is just circumstantial). The mean depth values give context clues to the surrounding environment, and this figure shows how the adaptability variables respond to them. Note that just for this figure, the displayed mean depth values from each sensor are always calculated using the maximum power levels to best and uniformly represent the context, even though different power levels are likely used as input into the auxiliary network, $h_\psi$, during evaluation.}
\label{bins}
\end{figure*}

\begin{figure}[htbp]
\centerline{\includegraphics{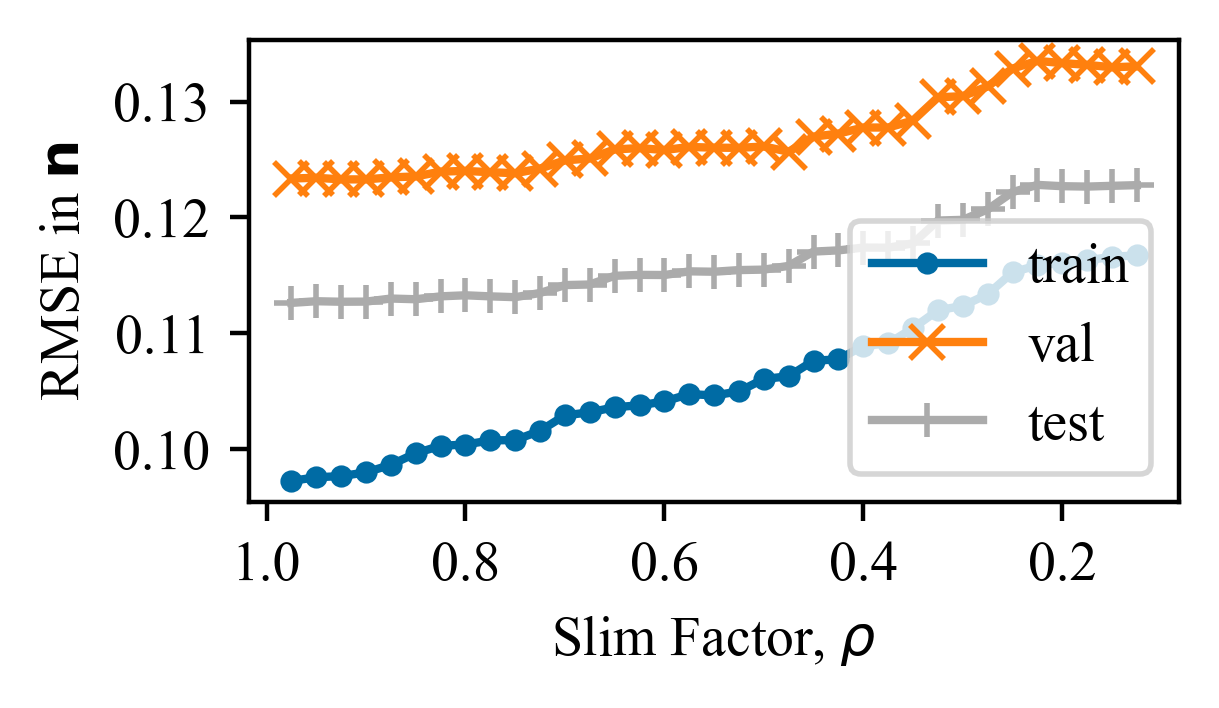}}
\caption{Results after reducing the resources allocated to computing a trained navigation model, $g_\phi$.  Shown is the Root Mean Squared Error (RMSE), calculated between the length-optimal navigation motions found from A-star and those predicted from $g_\phi$, as a function of the slimming factor, $\rho$, used to control the number of active parameters, $m$, in $g_\phi$. Not shown here, is that $m$ exhibits a quadratic decrease with $\rho$.}
\label{slims}
\end{figure}

Next we still evaluate the RMSE between length-optimal motions as found from A-star, $\mathbf{n}$, and those predicted from $g_\phi$, $\hat{\mathbf{n}}$; however, we now select one navigation model (using the seed with best RMSE) and evaluate how the error changes with varying values of the slimming factor, $\rho$. Fig.~\ref{slims} shows navigation RMSE as a function $\rho$ for each of the training, validation, and testing sets. This shows how the global RMSE increases with decreased $\rho$, warranting that if we use a static sub-network within the navigation super-network we would receive sub-par accuracy. The novelty of our approach is to intelligently and dynamically select the value of $\rho$, given context, as to not detrimentally decrease navigation accuracy -- which we experiment with next by training and evaluating the auxiliary network, $h_\psi$.


First we train the auxiliary network, $h_\psi$, for \emph{NaviSlim-C} which predicts the slimming factor, $\rho$, that directly controls the computing resources required to execute the navigation network. We evaluate \emph{NaviSlim-C} on three scenarios with increasing difficulty: 1) the Blocks map with only horizontal motion allowed, 2) the Blocks map with vertical motion unlocked, and 3) the City map with vertical motion locked. We compare these scenarios as we hypothesize that the computing resources, represented by $\rho$, required to run the navigation network will increase with increased difficulty. This hypothesize is warranted with the evidence provided in Table~\ref{tbl:resrho} which shows the average adaptation values used to reduce resource allocation, as learned by \emph{NaviSlim}, and evaluated on the test set. Listed are three scenarios with increasing difficulty. Where the "\textbf{Scen.}" column is the scenario being tested for, and the "$\eta$" column shows the relative resource expenditure reduction either for the number of active navigation network parameters, $m$, or sensor power level, $w$. All values shown in Table~\ref{tbl:resrho} are calculated using the test set. We see a significantly higher mean value of $\rho$ between each of these scenarios. The $\eta$ values in Table~\ref{tbl:resrho} are percent decreases over the larger super-network. Note that the numbers reported in Table~\ref{tbl:resrho} are for all evaluations done at the end of each curriculum learning step before increasing the distance between spawn and goal in Algorithm~\ref{alg:td3}, and all evaluation paths successfully reached their target position -- thus navigation accuracy is maintained.

\begin{table}[htbp]
\caption{}
\label{sensors}
\begin{center}
\begin{tabular}{|c|c|c|c|c|c|c|c|}
\hline
\textbf{Scen.} & \textbf{Map} & \textbf{Motion} & $\bar{\rho}$ & $\eta(m)$ & $\bar{p_f}$ & $\bar{p_d}$ & $\eta(w)$ \\ 
\hline
(1) & Blocks & Horizontal & 0.61 & 57\% & - & - & - \\
(2) & Blocks & Vertical & 0.88 & 86\% & 2.6 & 2.2 & 80\% \\
(3) & City & Horizontal & 0.93 & 92\% & 2.9 & 0.73 & 61\% \\
\hline
\end{tabular}
\label{tbl:resrho}
\end{center}
\end{table}

\begin{figure}[htbp]
\centerline{\includegraphics{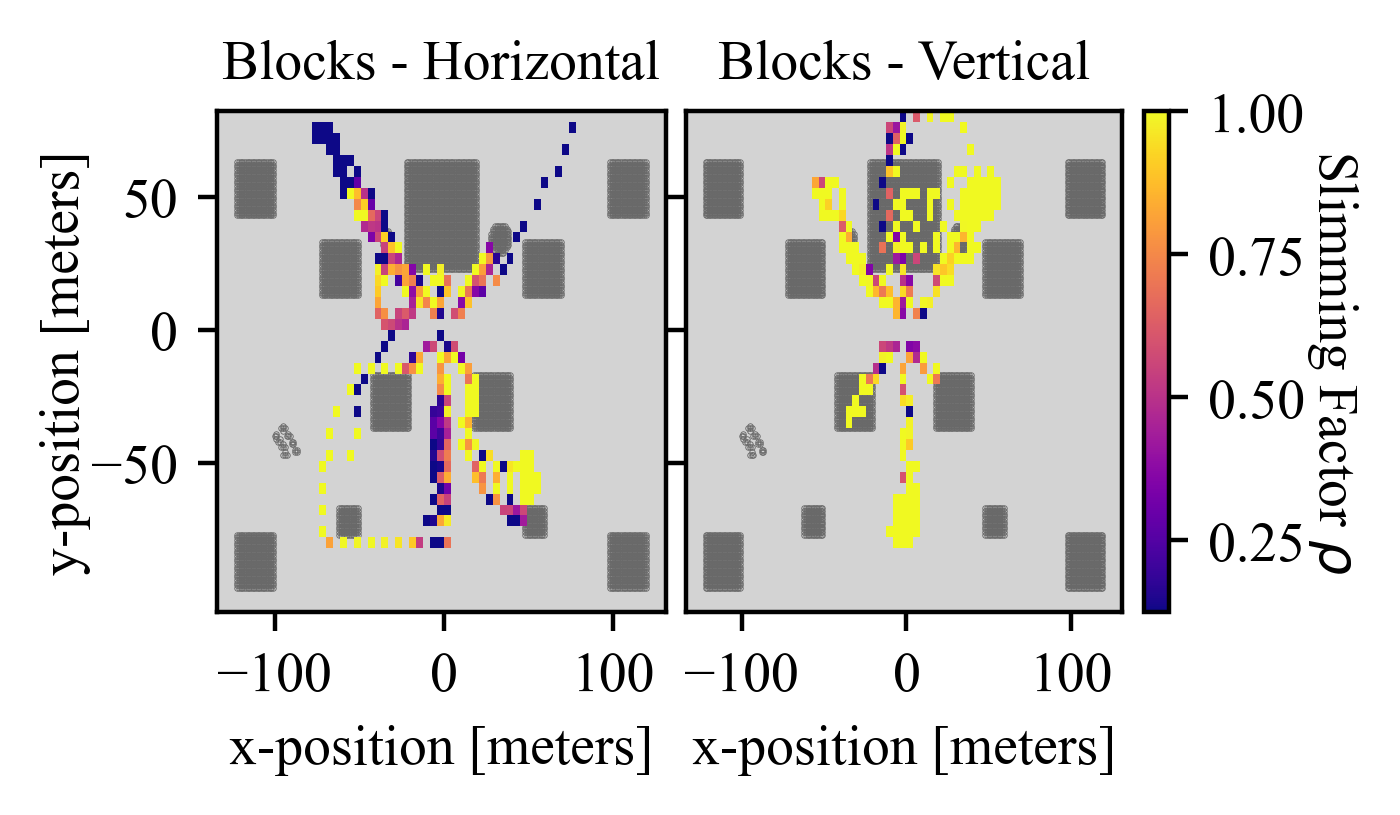}}
\caption{An aerial 2D view of the AirSim Blocks map overlaid with the average slimming factor, $\rho$, predicted from the auxiliary network, $h_\psi$, at each position. Included in the mean are only values from successful paths, and only those from the test set. The left panel shows a scenario that we consider where the drone can only move horizontally, and the right panel shows another scenario where vertical motion is unlocked. The darker gray shapes indicate objects the drone can collide with, however note on the right panel the drone is also flying over these objects.}
\label{heats1}
\end{figure}

Fig.~\ref{heats1} shows the learned average values of $\rho$ at each position in the Blocks map, using the test set. This figure is useful to see the dynamic nature of \emph{NaviSlim} and its context-aware behavior based on both the surrounding environment and maneuvers required to avoid collision with an object. We further isolate the context clues versus learned behavior as illustrated in Fig.~\ref{bins}, where we show the mean depth captured by each sensor versus $\rho$. Generally, when the navigation path is more convoluted there is a correspondingly high $\rho$, which is expected behavior for more complex navigation and maneuvers. 

Interestingly, as the distance between the drone and objects increases, so does the value predicted for $\rho$ from $h_\psi$. A possible explanation is that when objects are close, higher-level logic is not needed as the subspace of possible motions, that can be executed without colliding with that nearby object, shrinks. Similarly, after about 50 meters, $\rho$ begins to decrease with further increasing measured depth, likely because the environment becomes more open (less nearby objects and more open physical space) and the subspace of possible motions shrinks as more sophisticated maneuvers are not needed.

\begin{figure}[htbp]
\centerline{\includegraphics{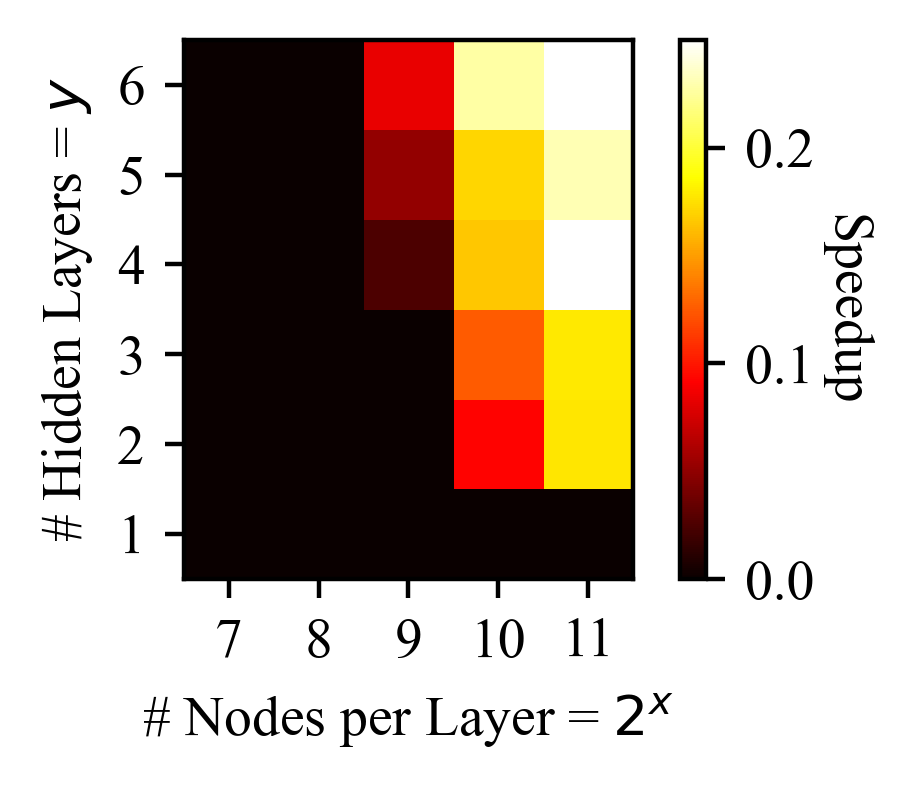}}
\caption{Test set results ran on a Jetson Nano to measure the relative speedups between using and not using \emph{NaviSlim}, as a function of navigation network size -- while using a fixed auxiliary network that has 2 hidden layers with 32 nodes in each. A speedup of zero (black grid spaces) indicates that using \emph{NaviSlim} takes more time to run than not using it, which is expected for smaller navigation network sizes due to the overhead of the auxiliary network.}
\label{nano}
\end{figure}

Next, we measure the actual differences in the resource cost (time, power, and energy) between using and not using \emph{NaviSlim} on a microprocessor similar to that typically deployed on micro-drones. We use a Jetson Nano with a Quad-core ARM Cortex-A57 MPCore processor and 4 GB 64-bit LPDDR4 1600MHz 25.6 GB/s memory. We compare relative resource costs by passing a static set of observations through: (1) \emph{NaviSlim}, including both the auxiliary and navigation modules, with the learned values for $\rho$; then (2) only the navigation network but with $\rho=1$ (\textit{i.e.}, just the static super-network without the auxiliary network). We measure the ratio difference between each resource as $\frac{v-u}{v}$, where \textit{u} is the resource cost associated with (1) \emph{NaviSlim} and \textit{v} is the resource cost associated with (2) the super-network. The size of the auxiliary hidden layers was fixed at [32, 32], while the size of the navigation hidden layers was varied - as shown in Fig.~\ref{nano} which shows the relative speedup associated with \emph{NaviSlim}.

We find that the difference in power consumption is negligible, with a mean relative difference of 0.005 and standard deviation of 0.07. Since energy consumption is power multiplied by time, this shows that execution time is the dominating factor in energy consumption. Note that reducing execution time of the navigation network also improves reaction time. Fig.~\ref{nano} shows that smaller networks can actually result in increased execution time when using \emph{NaviSlim} -- as indicated by the lower left corner with black grid spaces. This behavior is expected, since the overhead of the auxiliary network does not justify the small size of the navigation network. However, larger networks result in decreased run times - as indicated by the upper right corner with non-black grid spaces. This region is characterized by a positive speedup and also overlaps with the region with lowest navigation error as shown in Fig~\ref{hyper}. This proves that we can mitigate the larger execution times inherent with larger neural networks, which is needed to achieve lower navigation error, by using \emph{NaviSlim} -- noting that this speedup increases with size of the navigation network. 

Next we evaluate \emph{NaviSlim-S} which dynamically controls $p_f$ and $p_d$, the power levels of the forward facing depth sensor and downward facing depth sensor, respectively. Table~\ref{tbl:resrho} lists the average resolution levels for scenarios (2) and (3) after evaluating the trained auxiliary model, $h_\psi$, with the test set. The mean value of $p_f$ is greater than $p_d$ for each scenario, which is intuitive because most drone maneuvers involve moving horizontally rather than vertically. When using only horizontal motion, the downward facing depth sensor is almost completely turned off, with a mean $p_d$ value of 0.73. 

\begin{figure}[htbp]
\centerline{\includegraphics{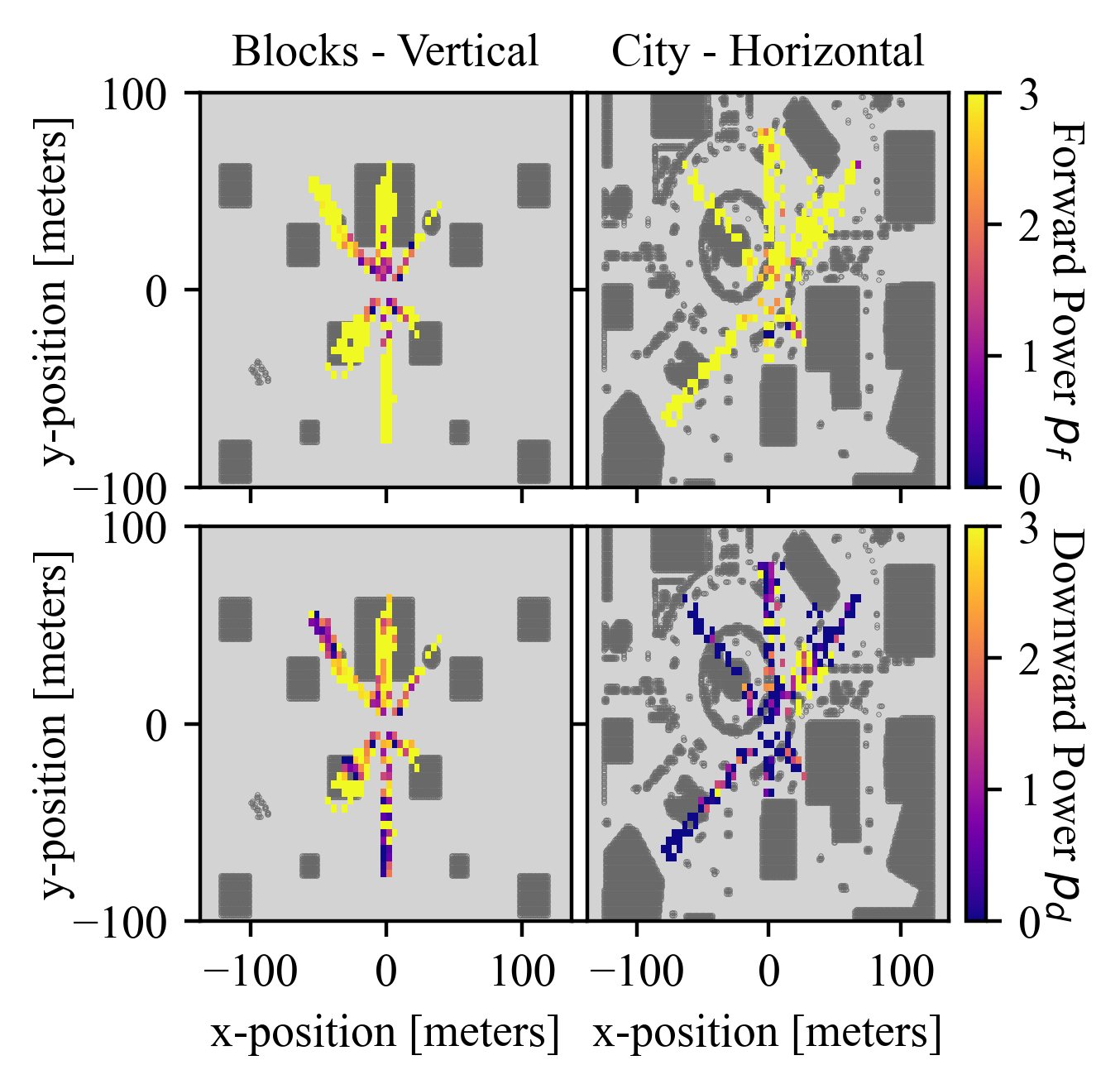}}
\caption{An aerial 2D view of the two AirSim maps overlaid with the respective power levels, $p_f$ and $p_d$, corresponding to the forward and downward facing depth sensors as predicted from the auxiliary network, $h_\psi$, at each position. Included in the calculations of each mean value are only those from successful paths in the test set. The left panels show results on the Blocks map with vertical motion unlocked. The right panels show results on the City map with vertical motion locked. The darker gray shapes indicate objects, however note that on the left panels the drone is also flying over these objects and some objects on the right panels are moving with time.}
\label{heats2}
\end{figure}

Fig.~\ref{heats2} and Fig.~\ref{bins} show the learned sensor power levels between the two scenarios as a function of context. From Fig.~\ref{bins}, we see that $p_f$ is independent of the downward depth sensor observations, but has a clear dependence on the forward depth sensor observations -- which is most intuitive. Interestingly, as the values returned from the forward depth sensor increase (indicating forward facing objects are further from the drone) so does $p_f$, which is a similar relationship we earlier observed with $\rho$ -- warranting that less resources are required when an object(s) is very near the drone. We observe another intuitive relationship that $p_d$ increases as the values returned from the downward depth sensor increase (indicating the drone is relatively higher in the air than objects below it). This relationship holds until the mean downward depth reaches some critical point at which $p_d$ then decreases.

\section{Conclusions}
\label{sec:conclusions}

We presented \emph{NaviSlim}, the first of its kind to dynamically scale computing and sensing used by a neural model for navigation of a (micro-)drone with extreme resource constraints. We detailed the training procedure used to obtain successful models that can safely navigate between points A and B, while using variable computing and sensing. We showed that an auxiliary neural network can successfully learn to map context to computing and sensing required by the difficulty of the current scenario. This is a novel evolution over static networks that must match computing and sensing of that required by the most difficult scenario. We showed that when deploying \emph{NaviSlim} to our test bed environment interfaced with the drone simulation tool Microsoft Airsim, we reduced average navigation model complexity between 57\% and 82\%, and sensing power levels between 61\% and 80\%, as compared to that of the static navigation network required to fulfill the same objectives. We posit that such methods will pave the way in a new evolution of dynamic neural networks used in resource constrained environments.

\section*{Acknowledgment}
This work was partially supported by the National Science Foundation under grants CCF 2140154, CNS 2134567, and DUE 1930546.

\bibliographystyle{IEEEtran}
\bibliography{bibliography}

\end{document}